\pdfoutput=1
\documentclass[11pt,table]{article}

\usepackage[final]{acl}

\usepackage{times}
\usepackage{latexsym}
\usepackage{url}
\usepackage[T1]{fontenc}
\usepackage[utf8]{inputenc}

\usepackage{microtype}
\usepackage{booktabs}
\usepackage{inconsolata}

\usepackage{graphicx}
\usepackage{paralist}
\title{UoB-NLP at SemEval-2025 Task 11: Leveraging Adapters for Multilingual and Cross-Lingual Emotion Detection}

\author{Frances Laureano De Leon  \and Yixiao Wang  \and Yue Feng \and Mark G. Lee \\
        University of Birmingham \\
        Birmingham B15 2TT}

\begin{document}
\maketitle
\begin{abstract}
Emotion detection in natural language processing is a challenging task due to the complexity of human emotions and linguistic diversity. 
While significant progress has been made in high-resource languages, emotion detection in low-resource languages remains underexplored.
In this work, we address multilingual and cross-lingual emotion detection by leveraging adapter-based fine-tuning with multilingual pre-trained language models. 
Adapters introduce a small number of trainable parameters while keeping the pre-trained model weights fixed, offering a parameter-efficient approach to adaptation.
We experiment with different adapter tuning strategies, including task-only adapters, target-language-ready task adapters, and language-family-based adapters.
Our results show that target-language-ready task adapters achieve the best overall performance, particularly for low-resource African languages with our team ranking 7th for Tigrinya, and 8th for Kinyarwanda in Track A. 
In Track C, our system ranked 3rd for Amharic, and 4th for Oromo, Tigrinya, Kinyarwanda, Hausa, and Igbo. 
Our approach outperforms large language models in 11 languages and matches their performance in four others, despite our models having significantly fewer parameters. 
Furthermore, we find that adapter-based models retain cross-linguistic transfer capabilities while requiring fewer computational resources compared to full fine-tuning for each language.

\end{abstract}

\section{Introduction}

Emotion detection in Natural Language Processing (NLP) remains a challenging task due to the complexity and nuance of human emotions. Language is often used in subtle and intricate ways to express emotion~\cite{Wiebe2005AnnotatingLanguage, Mohammad2018UnderstandingCategories} making emotion detection difficult not only for humans but also for state-of-the-art machine learning models. Accurately identifying emotions in text has broad applications across various fields, including commerce, public health, disaster response, and policymaking~\cite{semeval2018}, making continued research in this area essential. 
Emotion detection can take different forms, such as identifying the emotions of the speaker, determining what emotion a piece of text conveys, or detecting emotions evoked in a reader~\cite{Mohammad2023BestLexicons}. 
While significant progress has been made in high-resource languages like English and Spanish, emotion detection in low-resource languages remains underexplored.

This work focuses  on multilingual and cross-lingual emotion detection, in which the goal is to predict the perceived emotions of a speaker based on a given sentence or short text snippet.
This task presents several challenges. 
First, its multilingual and cross-lingual nature complicates emotion detection due to linguistic diversity. 
This diversity arises not only from the wide range of language families included in the datasets we use, but also from the varied sources used to compile the datasets~\cite{Muhammad2025BRIGHTER:Languages}.
Secondly, emotions are subjective and influenced by cultural norms, making it difficult to standardise emotion labels across languages~\cite{Woensel2019}.
Thirdly, multiple emotions can coexist within a single text, requiring models to handle multi-label classification effectively.

To address these challenges, we leverage adapter-based fine-tuning and multilingual language models. 
Adapters are parameter-efficient and modular components that introduce a small number of trainable parameters while keeping the pre-trained model weights fixed~\cite{Houlsby2019}. 
We investigate different adapter tuning strategies, incorporating both multilingual and cross-lingual approaches to improve model performance.
Our results indicate that target-language-ready task adapters outperform standard fine-tuning methods. Additionally, using adapters with multilingual Pre-trained Language Models (PLMs) yields better performance than Large Language Models (LLMs) for many of the languages~\cite{Muhammad2025BRIGHTER:Languages}. 
However, we also recognise challenges in applying the adapter approach to low-resource languages, where PLMs have limited pretraining data, and note its constraints in higher-resource languages.

Our contributions include:
\begin{compactitem}
    \item Evaluating different adapter-based approaches for multilingual and cross-lingual emotion detection.
    \item Investigating the impact of language-family-based adapter tuning on model performance.
\end{compactitem}

Our code and trained adapters will be publicly available on GitHub to support further research in this area~\footnote{\url{https://github.com/francesita/Adapters-EmoDetection-SemEval2025}}

\section{Related Literature}
As text-based communication continues to grow, extracting emotions from text has become a key area of research~\cite{emotion_survey_2024}. 
While sentiment analysis has been widely studied, it primarily categorises text into broad classes such as positive, negative, and neutral sentiment~\cite{Muhammad2025BRIGHTER:Languages}. 
In contrast, emotion detection aims to identify finer-grained emotions such as anger, fear, joy, and sadness, aligning with discrete emotion models like Ekman’s six basic emotions~\cite{Ekman1992AnEmotions} and Plutchik’s Wheel of Emotions~\cite{Plutchik1980}.
Most emotion detection research has been conducted in high-resource languages such as English, Spanish, and Arabic~\cite{emotion_survey_2024, semeval2018}, with some studies exploring Hindi, Bangla, and code-mixed languages like Hindi-English, Punjabi-English, and Dravidian-English~\cite{emotion_survey_2024}. 
However, challenges remain, particularly for low-resource languages. 
These include limited dataset availability, the ambiguity of emotional boundaries (necessitating multilabel classification), and the difficulty of detecting emotions from text alone~\cite{emotion_survey_2023}.

To address these challenges, we leverage adapters—lightweight modules that allow for efficient fine-tuning of language models without modifying their original parameters~\cite{Houlsby2019}. 
Adapters are inserted within the layers of a frozen model and trained to learn task-specific representations, making them effective for low-resource and multilingual settings~\cite{Houlsby2019, Pfeiffer2020MAD-X:Transfer}.
Previous work has shown that multilingual transformer models such as mBERT and XLM-RoBERTa perform poorly in low-resource languages, but adapters help mitigate these limitations~\cite{Pfeiffer2020MAD-X:Transfer}.
The standard adapter paradigm involves training Language Adapters (LAs) on unlabelled data with a Masked Language Modelling (MLM) objective, while Task Adapters (TAs) are trained on labelled task-specific data in a source language~\cite{Pfeiffer2020MAD-X:Transfer, Parovic2023}. 
For cross-lingual transfer, the task adapter is trained on top of a frozen source language adapter and later used for inference by swapping in a target language adapter. 
This approach enhances transferability but does not fully bridge the performance gaps in low-resource languages.
Recent studies have introduced target-language-ready task adapters, which train the task adapter by cycling through multiple language adapters, improving generalisation across languages~\cite{Parovic2023}. 
Another approach, phylogeny-inspired adapter training, incorporates linguistic relationships by structuring adapters according to language families, improving cross-lingual transfer~\cite{Faisal2022Phylogeny-InspiredLanguages}. Phylogeny-based adapter tuning has also been applied to multilingual sentiment analysis in African languages~\cite{gmnlp}.
Building on these approaches, we explore whether combining target-language-ready task adapters with the idea of training the adapters per language family, enhances emotion detection across diverse languages. 
We do this because languages within the same family often share structural and lexical similarities, potentially making cross-lingual transfer more effective compared to languages from different families~\cite{Faisal2022Phylogeny-InspiredLanguages}. 
Based on this, we train task adapters using language-family groupings to leverage these shared features, aiming to enhance knowledge transfer and improve performance within related languages.
We evaluate our models on the BRIGHTER~\cite{Muhammad2025BRIGHTER:Languages} and EthioEmo~\cite{Belay2025EvaluatingUnderstanding} datasets, which spans a broad set of languages, including many that are underrepresented in NLP research~\cite{Muhammad2025BRIGHTER:Languages}.

\section{Experimental Setup}
We conduct experiments using the SemEval-2025 Task 11~\cite{task_11} emotion detection datasets~\cite{Muhammad2025BRIGHTER:Languages, Belay2025EvaluatingUnderstanding}. 
The datasets have predefined training, development, and test splits.
The development sets are used for hyperparameter tuning and model selection, while the test sets are used for the final evaluation. 
The development sets are not used for training final models or adapters.  
SemEval Task 11 consists of three tracks, of which we participate in two: Track A, which focuses on multi-label emotion detection, and Track C, which involves cross-lingual emotion detection.
In both tracks, the objective is to determine whether each of six emotions—joy, sadness, fear, anger, surprise, and disgust—is present in the text. 
(For English and Afrikaans there are only five emotions.)
All experiments are conducted using xlm-roberta-base and afro-xlmr-base.  

\begin{figure}
    \centering
    \includegraphics[width=1.0\linewidth]{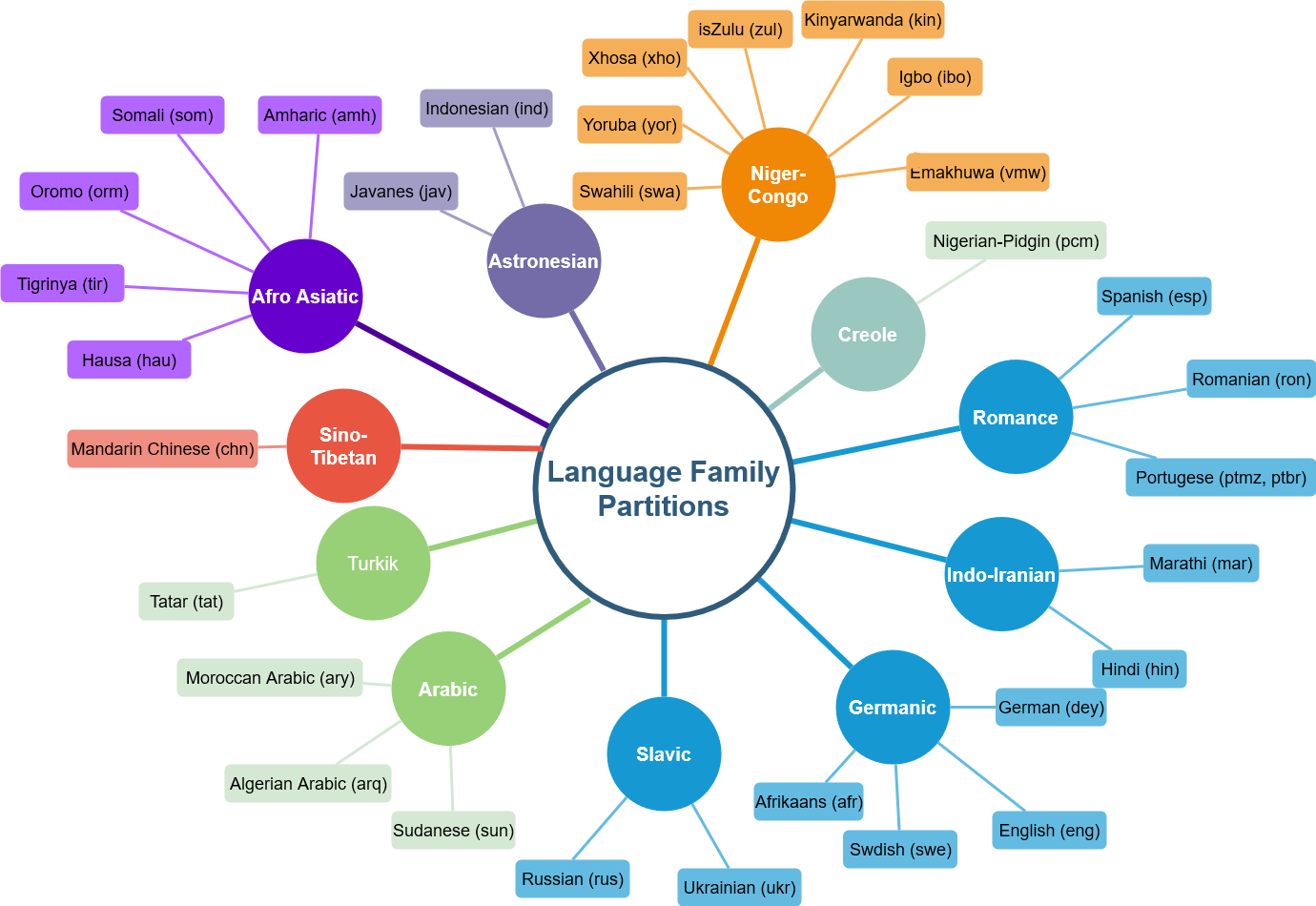}
    \caption{Languages used in this study and their partitioning into language families for adapter training.}
    \label{fig:lang-families}
\end{figure}

\begin{figure}
    \centering
    \includegraphics[width=0.9\linewidth]{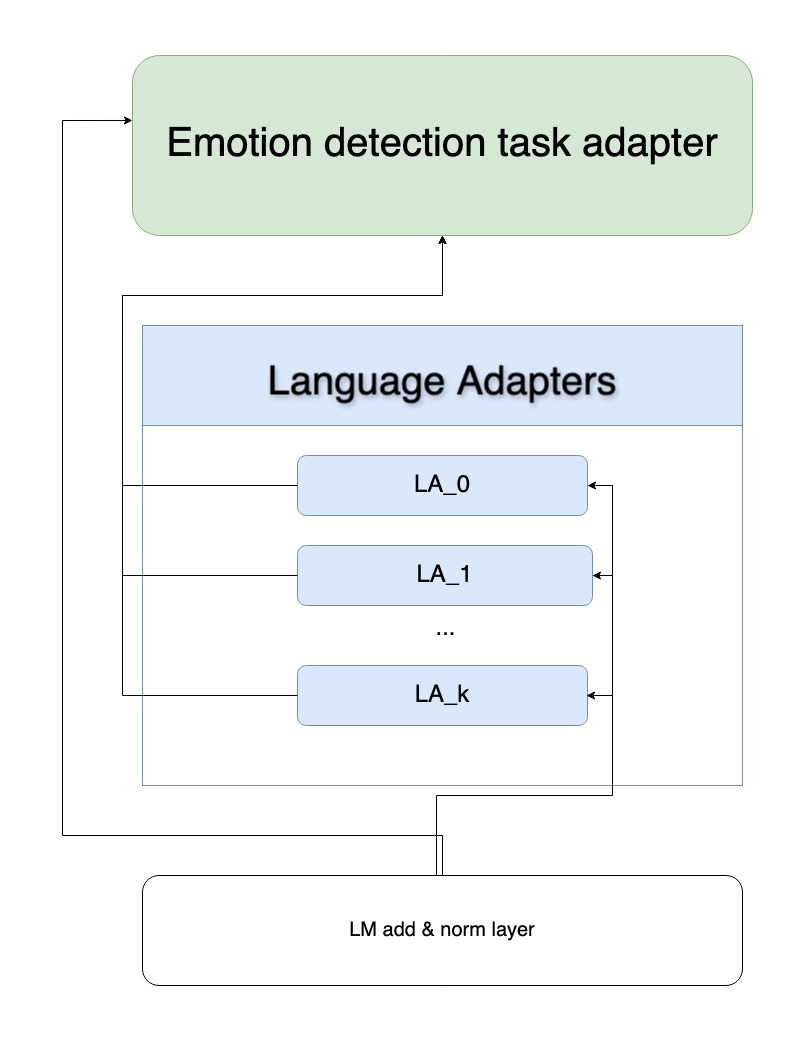}
    \caption{Illustration of a multilingual task adapter (TA) with a target language-ready (TLR) module at a single PLM layer. The diagram shows language adapters (LAs) for K target languages alongside the emotion detection TA. This method is from~\citet{Parovic2023}.}
    % Training is performed by cycling through the K LAs corresponding to the K languages.
    \label{fig:cycle_adapters}
\end{figure}

The organisers provide a baseline which we use as a reference. 
Additionally, we establish our own baseline by fine-tuning xlm-roberta-base and afro-xlmr-base individually for each language. 
These baselines serve as comparison points, and their results are presented in Table~\ref{tab:track_a}.  
We perform four main experiments. 
The first experiment follows the original adapter setup, where a task adapter is trained stacked on an LA, where the parameters are frozen.
The source language selected for this setup is Spanish.
We choose Spanish, rather than English, because it is both one of the most high-resource languages in the dataset, and it contains all 6 emotion labels. 
Additionally, we utilise the SemEval-2018 Task 1 dataset in Spanish to augment the emotion data for this experiment, ensuring the training of a robust TA.
During inference, the TA trained on Spanish emotion data is stacked on a LA corresponding to the target language. 
The second experiment investigates a task-only setup, where task adapters are trained for emotion detection without LAs. 
The third experiment involves training target-language-ready task adapters (TLR), following the approach of~\citet{Parovic2023}, in which a task adapter is trained using all available LAs corresponding to all the languages in the dataset. 
The fourth experiment focuses on language-family target-language-ready task adapters (lang-fam TLR), where TAs are trained using the previously described TLR cycling method,  but only cycling through the LAs of languages within the same family.
Figure~\ref{fig:lang-families} provides an overview of the language families considered in this work.  
Figure~\ref{fig:cycle_adapters} illustrates the adapter setup within the transformer, which is applied in the TLR, and lang-fam TLR adapter experiments. 
We use the HuggingFace Transformers~\footnote{\url{https://huggingface.co/}} and AdapterHub~\footnote{\url{https://adapterhub.ml/}}~\cite{Pfeiffer2020AdapterHub:Transformers} libraries for model training and evaluation.  

\section{Methodology}
Our approach consists of two key stages: LA training and TA training. 
We use pre-existing LAs for English, Spanish, German, Arabic, Chinese, Swahili, Hindi, and Russian, all available through the AdapterHub library~\cite{Pfeiffer2020AdapterHub:Transformers}. 
For most other languages, we train new LAs, except for Romanian and Emakhuwa, due to resource limitations. 
In total, our experiments involve 27 LAs.  

LAs are trained using the OSCAR dataset~\cite{OrtizSuarez2020ALanguages, OrtizSuarez2019AsynchronousInfrastructures} for languages such as Portuguese and Amharic, while Wikipedia~\cite{FoundationWikimediaDownloads} is used for low-resource African languages. 
Training follows prior work, with each LA trained for 100 epochs or 100,000 steps. 
The batch size is set to 8, the learning rate to 1e-5, and the maximum sequence length to 256.
For low-resource languages, training is conducted for 30,000 steps.
We use a bottleneck size of 16 of the adapters.  
Task adapters are trained by stacking them on top of the LAs, while keeping the parameters of both the base model and the LA frozen during training. 
A bottleneck size of 16 is also used, consistent with prior work. 
For TLR TAs, training is conducted using all LAs relevant to the dataset and experimental setup.

\begin{table*}[]
\centering
\resizebox{\textwidth}{!}{
\rowcolors{2}{gray!10}{white}
\begin{tabular}{l|c|c|c|c|c|c|c|c|c}
\toprule
\rowcolor{gray!30}
\textbf{Language} &
  \textbf{Competition results} &
  \textbf{LA\_t and TA} & % New column placed here
  \textbf{Xlm\_r\_target\_ready\_TA} &
  \textbf{Afro\_xlm\_r\_target\_ready\_TA} &
  \textbf{Family\_target\_ready\_TA} &
  \textbf{Xlm\_r\_TA} &
  \textbf{Afro\_xlmr\_TA} &
  \textbf{Xlm\_r\_baseline} &
  \textbf{Afro\_xlmr\_baseline} \\
\midrule
afr  & 0.4495 & 0.2639 & 0.3874 & 0.3629 & \textbf{0.4495} & 0.3594 & 0.3877 & 0.2962 & 0.2962 \\
amh  & 0.67   & 0.3757 & 0.6181 & \textbf{0.67} & 0.5609 & 0.6272 & 0.543  & 0.522  & 0.5369 \\
arq  & 0.4691 & 0.211 & 0.4691 & 0.4406 & \textbf{0.482}  & 0.4475 & 0.4145 & 0.1721 & 0.1721 \\
ary  & 0.5148 & 0.306 & 0.4328 & 0.4244 & \textbf{0.5148} & 0.4186 & 0.4017 & 0.2958 & 0.2457 \\
chn  & 0.5692 & 0.209 & \textbf{0.5916} & 0.5692 & 0.5158 & 0.5894 & 0.4794 & 0.5126 & 0.4081 \\
deu  & 0.548  & 0.4019 & \textbf{0.6041} & 0.548  & 0.5937 & 0.5947 & 0.4587 & 0.4538 & 0.4538 \\
eng  & 0.6451 & 0.3953 & 0.6451 & 0.6236 & \textbf{0.6581} & 0.6526 & 0.6236 & 0.6327 & 0.6468 \\
esp  & 0.7291 & 0.7324 & 0.7291 & 0.6985 & \textbf{0.7465} & 0.7283 & 0.6441 & 0.7303 & 0.7016 \\
hau  & 0.6628 & 0.2297 & 0.5864 & \textbf{0.6628} & 0.6271 & 0.5916 & 0.5557 & 0.5169 & 0.5169 \\
hin  & 0.8423 & 0.4049 & \textbf{0.8423} & 0.8  & 0.8411 & 0.8353 & 0.7404 & 0.8233 & 0.8378 \\
ibo  & 0.4718 & 0.1826 & 0.4585 & 0.4842 & 0.4718 & \textbf{0.4854} & 0.4331 & 0.3688 & 0.3398 \\
kin  & 0.5159 & 0.125 & 0.3178 & \textbf{0.5159} & 0.4657 & 0.3185 & 0.4561 & 0.3288 & 0.3288 \\
mar  & 0.7996 & 0.0989 & 0.8384 & 0.7996 & \textbf{0.8416} & 0.8291 & 0.5886 & 0.7924 & 0.7988 \\
orm  & 0.5261 & 0.1986 & 0.438  & \textbf{0.5261} & 0.4912 & 0.4514 & 0.4268 & 0.3931 & 0.2755 \\
pcm  & 0.5152 & 0.2582 & 0.5274 & 0.513  & 0.5152 & \textbf{0.5357} & 0.4902 & 0.112  & 0.112 \\
ptbr & 0.4776 & 0.4467 & 0.4788 & 0.4398 & 0.4776 & \textbf{0.4797} & 0.3919 & 0.2185 & 0.3091 \\
ptmz & 0.4004 & 0.242 & 0.3944 & 0.3247 & \textbf{0.4004} & 0.3849 & 0.2616 & 0.0778 & 0.1845 \\
ron  & 0.6811 & 0.4576 & \textbf{0.6811} & 0.6576 & 0.6592 & 0.6671 & 0.6349  & 0.1897 & 0.1886 \\
rus  & n/a    & 0.518 & 0.8246 & 0.8057 & \textbf{0.8282} & 0.8256 & 0.7591 & 0.8133 & 0.8137 \\
som  & 0.445  & 0.0717 & 0.3355 & \textbf{0.445} & 0.3873 & 0.3695 & 0.2315 & 0.2744 & 0.2944 \\
sun  & 0.3646 & 0.231 & \textbf{0.3646} & 0.2965 & 0.3359 & 0.3636 & 0.289  & 0.1872 & 0.1872 \\
swa  & 0.2624 & 0.2377 & \textbf{0.2624} & 0.2197 & 0.1624 & 0.2187 & 0.1497 & 0.1582 & 0.1582 \\
swe  & 0.5215 & 0.2342 & \textbf{0.5215} & 0.4829 & 0.5086 & 0.5165 & 0.4774 & 0.4119 & 0.4039 \\
tat  & 0.6371 & 0.2929 & \textbf{0.6371} & 0.5257 & 0.2743 & 0.64    & 0.4078 & 0.3776 & 0.3959 \\
tir  & 0.5029 & 0.1129 & 0.4149 & \textbf{0.5029} & 0.4154 & 0.4446 & 0.216  & 0.3357 & 0.3357 \\
ukr  & 0.5841 & 0.253 & 0.5841 & 0.4983 & \textbf{0.5846} & 0.5776 & 0.4531 & 0.4307 & 0.4276 \\
vmw  & 0.0564 & 0.0652 & 0.0564 & 0.0336 & 0.031  & \textbf{0.0766} & 0.011  & 0      & 0 \\
yor  & 0.1931 & 0.0343 & 0.1446 & \textbf{0.1931} & 0.1153 & 0.1406 & 0.1599 & 0.0987 & 0.0989 \\
\hline
\end{tabular}
}
\caption{Track A results for all languages included in our experiments. Model names ending in target\_TA refer to target-language ready adapters. LA\_t refers to LA for some target language.}
\label{tab:track_a}
\end{table*}

\begin{table*}[]
\centering
\resizebox{\textwidth}{!}{
\rowcolors{2}{gray!10}{white}
\begin{tabular}{l|c|c|c|c|c|c|c|}
\toprule
\rowcolor{gray!30}
\textbf{Language} & \textbf{Competition results} & \textbf{New Column} & \textbf{Xlm\_r\_target\_ready\_TA} & \textbf{Afro\_xlm\_r\_target\_ready\_TA} & \textbf{Family\_target\_ready\_TA} & \textbf{Xlm\_r\_TA} & \textbf{Afro\_xlmr\_TA} \\
afr  & 0.3629 & 0.2639 & 0.3874 & 0.3629 & \textbf{0.4495} & 0.3594 & 0.3877 \\
amh  & 0.6272 & 0.3757 & 0.6181 & \textbf{0.67} & 0.5609 & 0.6272 & 0.543 \\
arq  & 0.4406 & 0.211 & 0.4691 & 0.4406 & \textbf{0.482} & 0.4475 & 0.4145 \\
ary  & 0.4244 & 0.306 & 0.4328 & 0.4244 & \textbf{0.5148} & 0.4186 & 0.4017 \\
chn  & 0.5692 & 0.209 & \textbf{0.5916} & 0.5692 & 0.5158 & 0.5894 & 0.4794 \\
deu  & 0.5543 & 0.4019 & \textbf{0.6041} & 0.548 & 0.5937 & 0.5947 & 0.4587 \\
eng  & 0.6451 & 0.3953 & 0.6451 & 0.6236 & \textbf{0.6581} & 0.6526 & 0.6236 \\
esp  & 0.7291 & - & 0.7291 & 0.6985 & \textbf{0.7465} & 0.7283 & 0.6441 \\
hau  & 0.6271 & 0.2297 & 0.5864 & \textbf{0.6628} & 0.6271 & 0.5916 & 0.5557 \\
hin  & 0.8 & 0.4049 & \textbf{0.8423} & 0.8 & 0.8411 & 0.8353 & 0.7404 \\
ibo  & 0.4842 & 0.1826 & 0.4585 & 0.4842 & 0.4718 & \textbf{0.4854} & 0.4331 \\
ind  & 0.3329 & 0.3887 & 0.405 & 0.3329 & \textbf{0.4354} & 0.4027 & 0.3082 \\
jav  & n/a & 0.2825 & 0.311 & 0.2607 & \textbf{0.3519} & 0.2766 & 0.2201 \\
kin  & 0.4657 & 0.125 & 0.3178 & \textbf{0.5159} & 0.4657 & 0.3185 & 0.4561 \\
mar  & 0.7996 & 0.0989 & 0.8384 & 0.7996 & \textbf{0.8416} & 0.8291 & 0.5886 \\
orm  & 0.4912 & 0.1986 & 0.438 & 0.5261 & \textbf{0.4912} & 0.4514 & 0.4268 \\
pcm  & 0.513 & 0.2582 & 0.5274 & 0.513 & 0.5152 & \textbf{0.5357} & 0.4902 \\
ptbr & 0.4398 & 0.4467 & 0.4788 & 0.4398 & 0.4776 & \textbf{0.4797} & 0.3919 \\
ptmz & 0.3247 & 0.242 & 0.3944 & 0.3247 & \textbf{0.4004} & 0.3849 & 0.2616 \\
ron  & 0.6811 & 0.4576 & \textbf{0.6811} & 0.6576 & 0.6580 & 0.6671 & 0.6349 \\
rus  & 0.8057 & 0.518 & 0.8246 & 0.8057 & \textbf{0.8282} & 0.8256 & 0.7591 \\
som  & 0.3873 & 0.0717 & 0.3355 & \textbf{0.445} & 0.3873 & 0.3695 & 0.2315 \\
sun  & 0.2965 & 0.231 & \textbf{0.3646} & 0.2965 & 0.3359 & 0.3636 & 0.289 \\
swa  & 0.2197 & 0.2377 & \textbf{0.2624} & 0.2197 & 0.1624 & 0.2187 & 0.1497 \\
swe  & 0.4829 & 0.2342 & \textbf{0.5215} & 0.4829 & 0.5086 & 0.5165 & 0.4774 \\
tat  & 0.6371 & 0.2929 & 0.6371 & 0.5257 & 0.2743 & \textbf{0.64} & 0.4078 \\
tir  & 0.4446 & 0.1129 & 0.4149 & \textbf{0.5029} & 0.4154 & 0.4446 & 0.216 \\
ukr  & 0.4983 & 0.253 & 0.5841 & 0.4983 & \textbf{0.5846} & 0.5776 & 0.4531 \\
vmw  & 0.0336 & 0.0652 & 0.0564 & 0.0336 & 0.031 & \textbf{0.0766} & 0.011 \\
xho  & n/a & \textbf{0.1254} & 0.0254 & 0.0825 & 0.0568 & 0.0397 & 0.048 \\
yor  & 0.1931 & 0.0343 & 0.1446 & \textbf{0.1931} & 0.1153 & 0.1406 & 0.1599 \\
zul  & n/a & 0.0882 & 0.037 & \textbf{0.1159} & 0.082 & 0.0441 & 0.08 \\
\hline
\end{tabular}
}
\caption{Track C results for all languages included in our experiments. Model names ending in target\_TA indicate target-language-ready adapters. We exclude the esp score because we trained the TA using the Spanish language LA.}
\label{tab:track_c}
\end{table*}

% Please add the following required packages to your document preamble:
% \usepackage{graphicx}
% Please add the following required packages to your document preamble:
% \usepackage{longtable}
% Note: It may be necessary to compile the document several times to get a multi-page table to line up properly
% Please add the following required packages to your document preamble:
% \usepackage{graphicx}
% Please add the following required packages to your document preamble:
% \usepackage{graphicx}
% Please add the following required packages to your document preamble:
% \usepackage{graphicx}
% \usepackage[table,xcdraw]{xcolor}
% Beamer presentation requires \usepackage{colortbl} instead of \usepackage[table,xcdraw]{xcolor}
% Please add the following required packages to your document preamble:
% \usepackage{graphicx}
% \usepackage[table,xcdraw]{xcolor}
% Beamer presentation requires \usepackage{colortbl} instead of \usepackage[table,xcdraw]{xcolor}
% Please add the following required packages to your document preamble:
% \usepackage{graphicx}

\section{Results}
The best-performing model for emotion detection in our experiments was the TLR-TA, using XLM-RoBERTa-base for non-African languages and Afro-XLMR-base for African languages (see Tables~\ref{tab:track_a} and~\ref{tab:track_c}).
% Please add the following required packages to your document preamble:
% \usepackage{graphicx}

However, performance varied across languages. 
In Track A, our system ranked 7th for Kinyarwanda, 8th for Tigrinya, 11th for Hausa, 12th for Amharic, and Oromo.
In Track C, it ranked 3rd for Amharic, and 4th for Oromo, Tigrinya, Kinyarwanda, Hausa, and Igbo. 
These results suggest that our approach was particularly effective for low-resource African languages.

While our system showed moderate overall performance, it outperformed LLMs in 11 languages under a few-shot setting. 
In four additional languages, it achieved performance on par with LLMs, despite being significantly smaller and more computationally efficient~\cite{Muhammad2025BRIGHTER:Languages}. 
LLMs are trained on large-scale web data and are considerably larger in size, but they continue to struggle in low-resource settings. This is likely due to the limited presence of many languages in online data, which affects the models' ability to generalise across diverse linguistic contexts. 
Our use of adapters enables efficient parameter sharing and supports cross-linguistic performance while requiring fewer computational resources compared to full fine-tuning for each language.
These results suggest that using adapters with smaller pre-trained language models remains a relevant and effective approach, particularly in low-resource scenarios.

The best model was selected based on development set performance during the competition.
However, when evaluated on the test set, some models performed better than expected, while others underperformed. 
This discrepancy may be due to overfitting to the development set, differences in data distribution between the development and test sets, or varying levels of pretraining exposure to certain languages.
Performance also varied depending on the type of adapter used. 
The task-only adapters performed best for Igbo, Pidgin, Tatar, and Emakhuwa, while the TLR task adapters achieved the highest scores for 14 languages.
The Family-based task adapters performed best for two Arabic languages, two Romance languages, and Slavic languages. 
In our experiments, the traditional TA stacked on an LA for the target language performed poorly, except in the case of Xhosa.
In all cases, using TLR adapters performed similarly to or better than fine-tuning a PLM per language while retaining multilingual transfer capabilities.

These results demonstrate that adapters provide a modular and efficient approach to multilingual NLP.
A single base model can be adapted for different tasks and languages by inserting lightweight modules, reducing the need for extensive retraining. 
The findings align with previous research, reinforcing adapters as a flexible and scalable alternative to full-model fine-tuning.
% We also conducted an error analysis to identify where the model struggled the most. We found that... (insert findings here).

% Please add the following required packages to your document preamble:
% \usepackage{graphicx}
% \usepackage[normalem]{ulem}
% \useunder{\uline}{\ul}{}

\section{Conclusion}
Adapters offer a parameter-efficient method for training models in multiple languages for a given task. 
They are lightweight and allow language models to be applied to different tasks and languages without altering the original model weights. 
This approach is effective for cross-lingual transfer, outperforming the baseline in 28 languages and achieving strong results in multilingual emotion detection across 16 languages.
Furthermore, using adapters with pre-trained language models continues to improve performance over LLMs in mid- and low-resource languages. 
Future research will examine the use of prompt tuning methods with LLMs to assess their potential for specialised tasks and low-resource languages.

\section{Ethical considerations}
Adapters enhance multilingual performance, but rely on pre-trained models that may introduce biases. These biases can affect specific languages, dialects, or groups, which requires further evaluation and mitigation. Although adapters are effective for medium- and low-resource languages, performance gaps persist compared to high-resource languages, requiring improvement without reinforcing digital inequalities. 
This study covers a limited set of languages, leaving many low-resource languages underrepresented.

\section*{Acknowledgments}

The computations described in this paper were performed using the University of Birmingham's BEAR Cloud service, which provides flexible resource for intensive computational work to the University's research community. See http://www.birmingham.ac.uk/bear  for more details.% Bibliography entries for the entire Anthology, followed by custom entries
%\bibliography{anthology,custom}
% Custom bibliography entries only
\bibliography{references}

\end{document}